%% file: nlp_for_cat_ts.tex
\documentclass[letterpaper]{article}
\pdfoutput=1
\usepackage{cite}
\usepackage{amsmath,amssymb,amsfonts}
\usepackage{algorithmic}
\usepackage{graphicx}
\usepackage{textcomp}
\usepackage{xcolor}
\usepackage{array}

\def\BibTeX{{\rm B\kern-.05em{\sc i\kern-.025em b}\kern-.08em
		T\kern-.1667em\lower.7ex\hbox{E}\kern-.125emX}}

\usepackage{times,multicol,float,epsfig,amsmath, amssymb, graphicx}
\usepackage{multirow,subfigure} %Not installed on 105
\usepackage{amsthm}
\usepackage{algorithm,algorithmic,amsfonts}
\usepackage{mathtools}
\usepackage{helvet}
\usepackage{courier}
\usepackage[left=1in,right=1in,top=1in,bottom=1in,footskip=0.25in]{geometry}
\newcommand{\RR}{\mathbb R}
\newcommand{\rank}{\operatorname{rank}}
\newcommand{\col}{\operatorname{col}}

\newtheorem{obs}{Observation}

\begin{document}
\bibliographystyle{amsplain}

\title{Conference Paper Title*\\
	{\footnotesize \textsuperscript{*}Note: Sub-titles are not captured in Xplore and
		should not be used}
	\thanks{Identify applicable funding agency here. If none, delete this.}
}

\title{NLP Based Anomaly Detection for Categorical Time Series}

\author{%
	\begin{tabular}{c} Matthew Horak \\ Lockheed Martin Space \\ matthew.horak@lmco.com \end{tabular} \and
	\begin{tabular}{c} Sowmya Chandrasekaran\\ Lockheed Martin Space \\ sowmya.s.chandrasekaran@lmco.com \end{tabular} \and
	\begin{tabular}{c} Giovanni Tobar\\ Lockheed Martin Space \\ giovanni.a.tobar@lmco.com \end{tabular} }
\date{\vspace{-5ex}}

\maketitle

\begin{abstract}

Identifying anomalies in large multi-dimensional time series is a crucial and difficult task across multiple domains.  Few methods exist in the literature that address this task when some of the variables are categorical in nature.  We formalize an analogy between categorical time series and classical Natural Language Processing and demonstrate the strength of this analogy for anomaly detection and root cause investigation by implementing and testing three different machine learning anomaly detection and root cause investigation models based upon it.

\end{abstract}


\section{Introduction}
\input{introduction}

\section{Analogy to NLP}\label{S:nlp_analogy}
\input{nlp}

\section{Sentence-based Anomaly Scores}\label{S:sentence_as}
\input{sentence_as}

\section{Singular Value Decomposition Model}\label{S:svd}
\input{svd_model}

\section{End-to-end Transformer Model}\label{S:inline}
\input{transformer}

\section{Stand-alone Dense Embeddings}\label{S:stand_alone}
\input{stand_alone}

\section{Testing Data Sets}\label{S:data}
\input{data_sets}

\section{Model Performance}\label{S:performance}
\input{performance}

\section{ Conclusion}\label{S:conc}
\input{conclusion}

\section*{Acknowledgment}  This work was performed in a cross-company collaboration with a team of researchers at the NEC corporation with a forthcoming publication~\cite{nec2022collab}.

\bibliography{reference}

\end{document}

%% file: introduction.tex
%INTRODUCTION

Current time series anomaly detection techniques excel at dealing with continuous data.  As more and more mixed-type data have been collected, the need to more thoroughly address the categorical components of time series has increased as well.  However, the tools available for anomaly detection and root cause investigation in large, multivariate time series containing categorical variables have not kept pace with the need.  Additionally, in most anomaly detection contexts with time series involving many variables, knowing which variables are behaving most anomalously when a model indicates an anomaly is a great benefit to the user, but such explainability is not present in most current models dealing with categorical time series.

\subsection{Related work}
Recently, a wide variety of machine learning methods have been brought to bear on forecasting and anomaly detection in numerical time series, including Arima models~\cite{nec_siat}, neural networks~\cite{su2019robust, hundman2018detecting} and statistical methods ~\cite{tan2011fast, keogh2005hot}.  These methods can often be made to accommodate categorical data through one-hot encoding, but in the authors' experience quality of models in this family rapidly degrades as the fraction of the series' variables that are categorical increases.  The field in which anomaly detection in categorical time series is most developed in is intrusion detection in network security and fraud detection~\cite{padhi2021tabular,gwadera2005reliable}.  The authors in~\cite{padhi2021tabular} utilize a transformer architecture for fraud detection inspired by an analogy between finite sequences of discrete variables and words in the domain of Natural Language Processing (NLP).
%Another recent approach to categorical time series involves the dynamic modeling of the frequencies of each category~\cite{nec2022collab}.

\subsection{Contributions of this Paper}\label{S:contributions}
In this paper, we formalize and extend the analogy of multivariate categorical time series to classical concepts in NLP in a way well-suited to anomaly detection in multivariate time series arising from telemetry streams.  We design, implement and test three anomaly detection and root cause investigation models based on this analogy.  One  benefit of our formalization over methods in the current literature is its explainability in terms of an easy ranking of the variables in order of their apparent contribution to a suspected anomaly.  A second benefit is that current categorical time series anomaly detection algorithms in the literature are focused on settings with relatively few sensors, and our framework enables construction of models managing large sensor sets, as demonstrated by our results in Section~\ref{S:performance}.

The remainder of the paper up to the conclusion is organized as follows.  In Section~\ref{S:nlp_analogy} we develop the NLP analogy for categorical time series.  In Sections~\ref{S:svd},~\ref{S:inline} and~\ref{S:stand_alone} we present and test three anomaly detection and root cause identification models based on this analogy.  Sections~\ref{S:data} and~\ref{S:performance} respectively contain a description of our test data and the models' performances on it.

%% file: nlp.tex
\begin{figure}%[h]
	\centering
	\includegraphics[height = 1.5in]{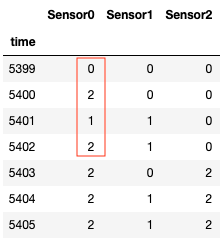}
	\caption{Sample Time Series}\label{F:series}
\end{figure}

In this section we go over the required modifications to concepts and terms of classical NLP and highlight each  in {\em italics} when it appears the first time. Concepts are illustrated with the time series segment shown in Figure~\ref{F:series}.  We refer to variables as {\em sensors}, the values reported by a sensor as {\em letters} and the set of letters said by a single sensor as that sensor's {\em alphabet}.

We regard the same value reported by two different sensors as different letters.  Thus, we regard all ``2's'' in the {\em Sensor0} column of Figure~\ref{F:series} as the same letter but different from the ``2's'' in the {\em Sensor2} column.  When distinguishing two different sensors' letters is necessary, we write, for example, ``Sensor0\_2'' and ``Sensor2\_2''.

Words are created by fixing a word length, say $wordL = 4,$ and defining a {\em word} as a sequence of $wordL$ consecutive letters that a sensor reports.  We index words by their last letter's position.  For example, the word that Sensor0 says at time $t = 5402$ consists of the sequence of the $4$ letters the sensor said during times $5399, 5400, 5401, 5402$, outlined in Figure~\ref{F:series}.

\begin{figure}%[h]
	\centering
	\includegraphics[height = 1.5in]{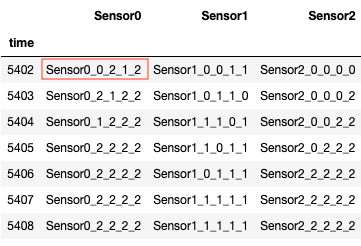}
	\caption{Words and Sentences}\label{F:df_sent_tr}
\end{figure}

A sensor's {\em vocabulary} is the set of words it says and the union of the vocabularies of all sensors of a time series is the series's {\em vocabulary}.   Since the alphabet of each sensor is unique to that sensor, every two different sensors' vocabularies are disjoint.  Therefore, we may shorten notation by removing the sensor name from each letter and placing it at the start of the word.  Thus, we write the word that Sensor0 says at time 5402 as ``Sensor0\_0\_2\_1\_2'', outlined in Figure~\ref{F:df_sent_tr}.

In order to capture inter-sensor relationships, we regard the set of words that all sensors in a time series say at a given time as a {\em sentence}.  Figure~\ref{F:df_sent_tr} shows the sentences constructed from the categorical time series in Figure~\ref{F:series}.  The sentence that the time series said at time $t$ is read across the time $t$ row of the dataframe.  Sensors of a time series generally have no naturally preferred order, so the words within a given sentence have no natural order unless we artificially order the sensors.  Therefore, the analogy to a natural language is not perfect, but our results show that we are able to retain enough temporal and inter-sensor information elsewhere to build highly effective models.

Finally, most applications in natural language processing require an ``out of vocabulary'' token for words that occur during inference but were never seen during training.  In this context, we find it best to include an unknown word token in every sensor's vocabulary, which we write as ``Sensor\_unknown\_word.''  It is also possible for a previously unseen {\em letter} to occur during inference, so we introduce a second unknown word token, ``Sensor\_unknown\_letter'', to each sensor's vocabulary and use Sensor\_unknown\_word when the unknown word involves only letters seen in training and Sensor\_unknown\_letter when the word involves a letter never seen in training.

%% file: sentence_as.tex
In the remainder of this paper, we describe and report the performance of three anomaly detection and root cause identification models built with the corpus of sentences resulting from processing a categorical time series into a sequence of sentences.  Given a time series $TS_{tr}$ of training data and a time series $TS$ (using the same sensors) in which we aim to detect anomalies, all three methods rely on computing an anomaly score $as(x_t)$ for the sentence $x_t$ that the inference series $TS$ says at time $t$, with higher score indicating more anomalous behavior.  To simplify the analysis here, we classify times as nominal or anomalous, by simply setting a {\em threshold factor} $\alpha$ and flagging as anomalous any time $t$ at which $as(x_t)$ exceeds $\alpha$ times the value of the $99.5^{th}$ percentile of the anomaly scores observed when inference is performed on the training data or a set of validation data.

\subsection{Root Cause Investigation}\label{S:root}
In the models we demonstrate, the score $as(x_t)$ is always a sum of word-based anomaly scores for each word in the sentence $x_t$.  Since every two sensors' vocabularies are disjoint, words are associated with unique sensors.  Also, every sentence contains exactly one word from every sensor's vocabulary.  Therefore, at any given time, sensors can be ranked according to the magnitude of the contribution of the words associated with them.  Additionally, for a set of (possibly non-consecutive) times $\{t_1, t_2, \ldots, t_M\}$ during which an anomaly is suspected, we define the {\em suspect score} of sensor $s$ over those times as the sum of its words' contribution to $as(x_t)$ over all times $t \in \{t_1, t_2, \ldots, t_M\}$.  We denote the sensor-specific score over these times as $score(s)$.  This score ranks sensors by their apparent contribution to a suspected anomaly over a time period.  This highly interpretable ranking offers a valuable degree explainability for users interested in root cause analysis.

\subsection{Next Sentence Forecast Models}\label{S:next_sentence}
For the model of Section~\ref{S:svd}, $as(x_t)$ is calculated only from $x_t$, but the models of Sections~\ref{S:inline} and~\ref{S:stand_alone} rely on a ``next sentence forecast'' framework.  For both of these models, we fix a lookback length, $N$.  At time $t$, the input is a sequence of consecutive sentences from the inference series $TS$, $(x_{t-N}, x_{t-N+1} \ldots, x_{t-1})$, and the output is a forecast, $\hat{x}_t$ for sentence $x_t$.  

The models are trained on nominal data, so  $\hat{x}_t$ is a forecast for $x_t$ under nominal conditions as characterized by the training sentences from $TS_{tr}$.  Therefore, in each model, a numerical measure of the difference between $\hat{x}_t$ and $x_t$ serves as the anomaly score $as(x_t)$.

%% file: svd_model.tex
In this section, we develop an anomaly detection model based on the Singular Value Decomposition (SVD) of the term-document matrix of $TS_{tr}$.  This decomposition has previously been used to uncover latent semantic structure in documents~\cite{deerwester1990indexing} and perform other NLP tasks~\cite{zweig2012computational}.

\subsection{Term Document Matrix}\label{S:svd_W}
Arbitrarily order the vocabulary $V = (w_1, w_2, \ldots, w_m)$ and the sentences $S = (s_1, s_2, \ldots, s_n)$ of the training corpus.  The {\em term frequency} of word $i$ in sentence $j$, $TF(i, j)$, is the number of times that $w_i$ occurs in $s_j$.  For the {\em inverse document frequency} of word $i$, we use the  formula  $IDF(i) = \log\frac{n+2}{f_i+1}$, where $n$ is the total number of sentences, and $f_i$ is the number of sentences containing word $i$.  The {\em term frequency matrix} for the corpus is the matrix $W$ whose entry in position $(i,j)$ is $W(i,j) = TF(i,j)\cdot IDF(i).$  This formula gives all unknown\_word and unknown\_letter tokens IDF weight $\log(n+2),$ which is larger than all other ``true'' words but can be modified as described  Section~\ref{S:performance}.  Column $j$ of $W$ is a kind of $m$ dimensional representation of the sentence $s_j$.

\subsection{Projection via SVD}
Next, fix a dimension $k\leq\rank(W)$ for a subspace onto which the sentences are to be projected and construct the full singular value decomposition, $$W = U_0 \Sigma_0 V_0^T.$$  Let $U$ be the $m \times k$ matrix consisting of the first $k$ columns of $U_0$, $\Sigma$ be the $k \times k$ matrix consisting of the first $k$ rows and columns of $\Sigma_0$ and $V$ be the $k \times n$ matrix consisting of the first $k$ rows of $V_0$.  There are two key observations.

\begin{obs}\label{O:eckart_young}
		If $W' = U \Sigma V^T,$ then $W'$ has rank $k$ and $W \approx W'$.  The approximation is optimal over all matrices of rank $\leq k$ in the sense that if $A$ is any $m\times n$ matrix with $\rank(A) \leq k$ then $\|W - W'\| \leq \|W - A\|.$
\end{obs}
\begin{obs}\label{O:proj}
	The matrix $U$ defines an orthogonal projection $P: \RR^m \to \RR^m$ of $\,\RR^m$ to itself by the formula
	\begin{equation}\label{E:p}
	P(x) = UU^Tx.
	\end{equation}
	The image of $P$ is the column space of $W', \col(W')$.
\end{obs}

Observation~\ref{O:eckart_young} is the Eckart-Young-Mirsky Theorem~\cite{eckart1936approximation}.  The proof of  Observation~\ref{O:proj} relies heavily on the fact that $U^TU  = I$ because $U_0$ is orthogonal, as well as the fact that $\dim(\col(W')) = k = \dim(\col(U))$.

\subsection{Anomaly Score and Root Cause}\label{S:svd_as}
Observations~\ref{O:eckart_young} and~\ref{O:proj} provide the optimal $k-$dimensional subspace to project onto in the sense of minimizing the average distance between $x_t$ and $P(x_t)$ for all columns of $W$.  The average is minimized not just over projections onto $\col(W')$ but over all possible projections onto all possible $k-$dimensional subapaces of $\RR^m$.  This justifies the interpretation of $P(x)$ as a kind of optimally in-family $k-$dimensional reconstruction of the $m-$dimensional sentence represented by $x$.  Therefore, for an arbitrary sentence, $x$, we define its anomaly score by,
\begin{equation}\label{E:ew}
as(x) = \|P(x) - x\|= \sum_{i=1}^{m}((P(x_t))_i - (x_t)_i)^2.
\end{equation}
The decomposition of $as(x)$ in terms of individual word contributions on the right of Equation~\ref{E:ew}  lets us define the individual sensor scores $score(s)$ for root cause investigation with the method described in Section~\ref{S:root}.

%% file: transformer.tex
\subsection{ Transformer Architecture}

The transformer model was first introduced by Vaswani et al.~\cite{vaswani2017attention} for the purpose of language translation using an encoder-decoder architecture. We repurpose the encoder portion of this model to perform both encoding and decoding in the same layer. For this conversion, the attention mechanism was modified to perform two types of attention: self-attention and causal attention.

\begin{figure}%[h]
	\centering
	\includegraphics[width=3in]{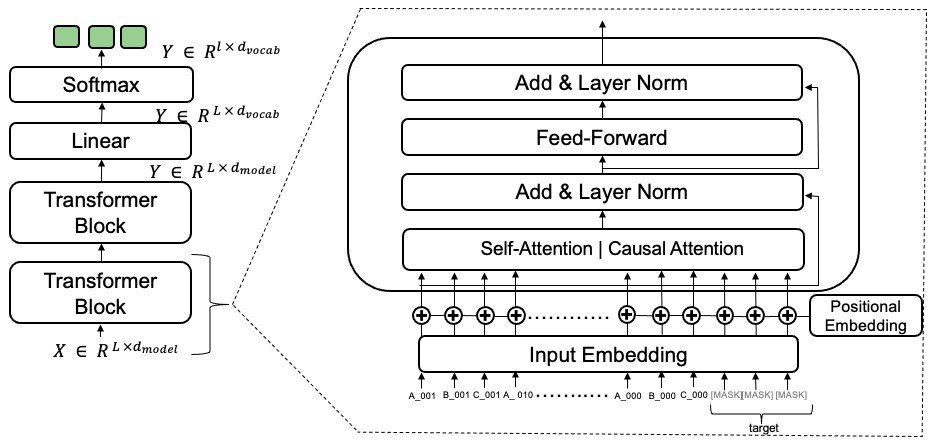}
	\caption{Transformer model.  $L$: sentence length, $l$: target length, $d_{model}$: embedding dimension, $d_{vocab}$: vocabulary size, $d_{sensor}$: sensor vocabulary size}\label{F:transformer}
\end{figure}

The right side of Figure~\ref{F:transformer} describes the architecture of each transformer block. From the bottom, we transform the sequence of sentences into an embedding vector which represents each word. The words are then added with another embedding vector called positional embedding. This embedding is the same size as the input embedding and it is designed to carry information about the position of each word of the sentence.  Once the data passes through the embedding layer, the transformer block splits the embedded data based on the number heads used for each block. The number of heads and the embedding dimension are tunable hyperparameters, constant for each transformer block. The number of transformer blocks in the full model is also a tunable hyperparameter.

\subsection{Attention Mechanism}

The attention mechanism shown in Figure~\ref{F:transformer} is composed of two different mechanisms: a self-attention mechanism and a causal attention mechanism. The self-attention is used only on the unmasked words to build a rich representation of the words by looking at future words, as well as previous words. On the other hand, the structure of the causal attention used on the masked words ensures no valuable information is transmitted from future masked words. This causal attention forces the model to build a representation of the masked words based only on the previous words. Researchers He et. al~\cite{he2018layer} have shown that this type of dual attention mechanism outperforms transformer baseline architecture in machine translation tasks. This architecture also reduces the number of hyperparameters by more than half by implementing both tasks in a single block.

\subsection {Sentence prediction}

As shown on the left side of Figure~\ref{F:transformer}, the output dimension of each transformer block is the same as that of the input embedding. This output embedding is then linearly transformed into the dimension of size of the vocabulary. The layer has a softmax activation because the target is the one-hot-encoding of each word in the input. The output is finally filtered only to pass positions of the masked sentence on to the cross-entropy loss, but this is not shown in the figure.  In practice, we compute the loss on each sensor’s vocabulary independently instead of computing the overall loss. We then average the sensor-specific losses to obtain the total loss. This technique comes from Padhi et. al [6], where researchers implement a global vocabulary and local vocabulary to compute the cross-entropy loss for each categorical column in their dataset.

\subsection{Anomaly Score an Root Cause}\label{S:tr_as}

The trained model generates next sentences based on nominal characteristics exhibited by the training data, so a measure of the difference between the forecast $\hat{x}_t$ and the actual $x_t$ is used as the anomaly score at time $t$.  For this, we used the Levenshtein distance, which is the number of changes required to convert the predicted word to the truth word.  The Levenshtein scores are then added together across words of the sentence to arrive at $as(x_t)$,  as shown in Equation~\ref{E:lev}, where $w$ is the actual word, $\hat{w}$ is the model's predicted word and $lev(w,\hat{w})$ is the Levenshtein distance between $w$ and $\hat{w}$. 
\begin{equation}\label{E:lev}
as(x_t) = \sum_{w \in x_t}lev(w,\hat{w}).
\end{equation}

Anomaly score $as(x_t)$ is used to flag anomalous times using the threshold method of Section~\ref{S:sentence_as}.  Since words are associated with unique sensors, Equation~\ref{E:lev} allows us define the individual sensor scores $score(s)$ for root cause investigation with the method described in Section~\ref{S:root}.

%% file: stand_alone.tex
In this section, we describe a method of creating dense vector representations, or embeddings, of the words in a time series's vocabulary, akin to Word2vec or GloVE representations in classical NLP.  We also show their use in a downstream task by demonstrating an anomaly detection model based on these representations.

\subsection{Entity Embedding}\label{S:ee}
 While existing NLP algorithms could be used to obtain vector representations of words arising from a categorical time series in an unsupervised manner, our situation is different from classical NLP because words in a sentence here have no naturally preferred ordering.  We work around this difference by using a trainable entity embedding layer placed before a deep neural network as introduced in~\cite{guo2016entity}.  The full model is trained on the task of masked word modeling, and the activations of the trained embedding layer are used as the vector representations.

\subsection{Implementation}\label{S:ee_implement}
\begin{figure*}[!t]
	\centering
	\includegraphics[width=5in]{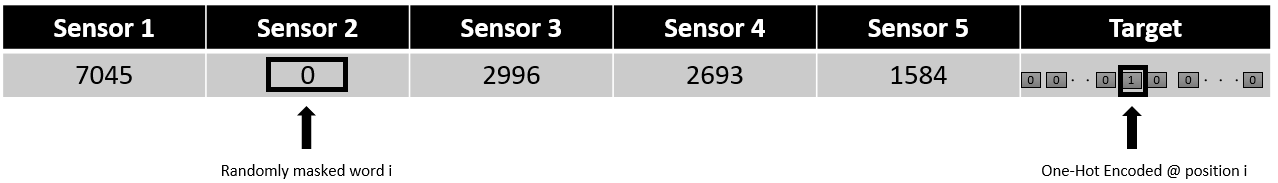}
	\caption{Masked sentence/target pair}\label{F:masking}
\end{figure*}

While more sophisticated architectures could certainly provide performance gains, for demonstration purposes, we used a simple three-layer feed-forward network after the embedding layer.  Each training sample consists of an input sentence with a randomly selected word masked to the value $0$, and the target is the one-hot encoding of the masked word.  As shown in Figure~\ref{F:masking}, the words in each sentence are first integer encoded with strictly positive integers so that the embedding can accommodate the $0$-value mask.  This has the further benefit that in downstream tasks, the vector representation corresponding to the value $0$ can be used for unknown word tokens' representations.  Once the model is fully trained, the learned parameter matrix of the embedding layer is used as a lookup table for the dense embeddings of the words.

\subsection{LSTM}\label{S:LSTM}

The vector representations of the previous section can be used to fuel most any machine learning model, but we again demonstrate their use on anomaly detection by next sentence forecasting described in Section~\ref{S:next_sentence}.  To demonstrate their ``out-of-the-box'' applicability to downstream tasks, we used a simple LSTM-based model with two LSTM layers followed by a dense layer with softmax activation to make the final forecasting.

The words in each sentence $x_i$ of an input sequence $(x_{t-N}, x_{t-N+1} \ldots, x_{t-1})$ are each embedded according to the embeddings learned above.  The resulting 2-dimensional representation of $x_i$ is flattened to $1$-dimensional representation $\overline{x}_i$ and the sequence $(\overline{x}_{t-N}, \overline{x}_{t-N+1} \ldots, \overline{x}_{t-1})$ is the model's input.  Each sentence has exactly one word for each sensor, so the target sentence $x_t$ can be encoded in a kind of ``multi-hot'' fashion, consisting of a vector of length equal to the total number of words in the full vocabulary with a ``1'' in the position of every word occurring in the target sentence.  The model is trained on the cross entropy loss.

\subsection{Anomaly Score and Root Cause}\label{S_ee_as}

For anomaly scoring during inference, we first translate the model's forecast at time $t$ back into sentence form.  Words in each sentence correspond exactly with the sensors of the time series, so as in Section~\ref{S:tr_as} we use the sum of the Levenshtein distances between the forecasted word for each sensor and the the actual word for that sensor observed in sentence $x_t$ as the anomaly score $as(x_t)$.  Since $as(x_t)$ is the sum of word-based contributions, we are able to define the individual sensor scores $score(s)$ for root cause investigation with the method described in Section~\ref{S:root}.

%% file: data_sets.tex
Authors in~\cite{hundman2018detecting} evaluate several anomaly detection algorithms on $82$ pairs of training/testing time series pairs originating form two assets managed by the NASA Jet Propulsion Laboratory.  They contain a single response sensor and $24$ binary command indicators.  We evaluate our models on the $7$ pairs in which the response sensor reports $5$ or fewer values, whose ordinal positions we take as the categories.  The train/test pairs used for this analysis are $E_1, E_5, E_6, E_{10}, E_{11}, E_{12}$ and $G_7$.  For testing series, we use the first $1000$ time steps to establish the ordinal ranking.  Aggregate $F_{0.5}$ scores over all data sets is reported in~\cite{hundman2018detecting}, with scores for the several models they evaluated ranging from a low of $0.44$ to a high of $0.71.$  We follow~\cite{hundman2018detecting} in focusing on models forecasting only the single response variable.

We also use a second collection of four multivariate categorical training/testing time series pairs originating from data streams internal to Lockheed Martin.  Each of the testing series has a single anomaly identified by the operator of the device generating the series.  These time series contain between $319$ and $655$ categorical sensors each, which report up to $5$ distinct values apiece.  The sensors of each time series are also grouped into $7$ ``engineering subsystems'' of sensors associated with different aspects of the device originating the data.

Counting false positives, false negatives and true positives is somewhat challenging due to the fact that we would like to regard each multi-time-step anomaly period as a single anomaly event.  We made these calls by hand on a case by case basis, using criteria established in work with subject matter experts (SMEs) who indicated that if a tool flags an anomaly within about $1\%$ of its start (relative to the length of the series) it should be classified as a ``successful'' identification, hence true positive.  If the tool did not flag any time period within about $1\%$ of the start, it is classified as a false negative.  Generally the times when models flag anomalies tended to cluster well, so we counted every cluster not overlapping an anomaly as a false positive.  Using these conventions, we report $F_1$ and $F_{0.5}$ for each model.

%% file: performance.tex
We found that a good word length for all models except the SVD models on the JPL data sets was $wordL = 5$.  For the SVD model on JPL data sets, $wordL = 20$ worked best, probably because the increased number of words helped compensate for the fairly small dimension that raw data.  For both data sets our {\em threshold factor} for flagging anomalies as described in Section~\ref{S:sentence_as} is $\alpha = 1.25$.  Hyperparameter choices specific to each specific model are as follows.

For the SVD model, setting the TFIDF value of the unknown word tokens to twice the maximal value of all ``true'' words seemed to balance well the importance of unknown letters with general sensitivity to anomalies involving known letters.

For the transformer model, the embedding dimension was $256$, the number of heads was $5$, and $2$ transformer blocks were used in the full model.

For the stand alone dense embedding and LSTM model, the embedding dimension for JPL data was  $2$ and for the Lockheed Martin data it was  $5$.  The output sizes of the first two layers in the masked word forecasting model were the number of sensors and the average of the number of words and the number of sensors.  The LSTM model had two LSTM layers, the first of dimension $N_1 = \max(N/2, 800)$ where $N$ is the number of sensors and the second of dimension equal to the average of $N_1$ and the total number of words.  The lookback window for both LSTM layers in both cases was $10$.

For the SVD and stand-alone dense embedding models, a separate model was built for each of the seven subsystems in the Lockheed Martin data, and an ensemble voting method was used to flag anomalous periods.  For the inline model, a single model was made on all of the binary sensors of each Lockheed Martin data set.  Performance of these models is shown in Table~\ref{T:svd_all}.

\begin{table}%[h!]
	\begin{center}
		\begin{tabular}{c||c|c|c|c|c|c|c|}
											   & $TP$  & $FP$ & $FN$    & $F_1$         & $F_{0.5}$ \\ \hline \hline
			SVD on JPL     				& $13$  & $6$   & $0$      & $0.81$      & $0.73$ \\ \hline 
			SVD on LM 	  				& $4$    & $2$   & $0$     & $0.80$      & $0.71$ \\ \hline
			Inline on JPL   			 & $9$  & $6$   & $4$     & $0.64$      & $0.62$ \\ \hline 
			Inline on LM  				 & $4$    & $0$   & $0$    & $1.0$      & $1.0$ \\ \hline
			 Stand Alone on JPL      & $12$  & $3$   & $1$    & $0.86$      & $0.82$ \\ \hline 
			Stand Alone on LM		& $4$    & $3$   & $0$    & $0.73$      & $0.63$ \\ \hline 
		\end{tabular}
		\caption{Performance on JPL and Lockheed Martin (LM) data}\label{T:svd_all}
	\end{center}
\end{table}

The JPL data involved only one response sensor, so the suspicious sensor ranking aspects of the models did not apply to that data.  For every anomaly in the Lockheed Martin data for which the SMEs identified categorical sensors involved in the root cause, all but one of our models placed more than $25\%$ of the SME-identified anomalous sensors in the top $10\%$ of suspect sensors.

%% file: conclusion.tex
This paper has three main contributions.  First, we formalize the analogy between categorical time series and texts in classical Natural Language Processing to a full framework in which NLP tools can be applied to general categorical time series with minimal modification.  Second, we show how the formalization of words and sentences in this framework enables model explainability useful in root cause investigations into anomalies in categorical telemetry streams.   Finally, we demonstrate the applicability of this framework to series with hundreds of categorical sensors by developing and testing three anomaly detection and root cause models within this framework.